%% file: paper.tex
\documentclass[]{bytedance_seed}

\usepackage[toc,page,header]{appendix}

\usepackage{scrextend}

\input{math_commands.tex}

\usepackage{hyperref}
\usepackage{url}
\usepackage{graphicx}
\usepackage{wrapfig}
\usepackage{booktabs}
\usepackage{tabularx}
\usepackage{multicol}
\usepackage{multirow}
\usepackage{makecell}
\usepackage[table]{xcolor}
\usepackage[utf8]{inputenc}
\usepackage[flushleft]{threeparttable}

\usepackage{minitoc}

\title{FSVideo: Fast Speed Video Diffusion Model in a Highly-Compressed Latent Space}

\author{FSVideo Team}

\affiliation{Intelligent Creation, ByteDance}
\abstract{
We introduce FSVideo, a fast speed transformer-based image-to-video (I2V) diffusion framework. We build our framework on the following key components: 1.) a new video autoencoder with highly-compressed latent space ($64\times64\times4$ spatial-temporal downsampling ratio), achieving competitive reconstruction quality; 2.) a diffusion transformer (DIT) architecture with a new \textsl{layer memory} design to enhance inter-layer information flow and context reuse within DIT, and 3.) a multi-resolution generation strategy via a few-step DIT upsampler to increase video fidelity. Our final model, which contains a 14B DIT base model and a 14B DIT upsampler, achieves competitive performance against other popular open-source models, while being \textbf{an order of magnitude} faster. We discuss our model design as well as training strategies in this report.
}

\date{\today}
\correspondence{Xiao Yang at \email{xiaoyang\_seravee@outlook.com}}
\projectpage{\url{https://kingofprank.github.io/fsvideo/}}

\begin{document}
\maketitle

\begin{figure}[h]
    \centering\vspace{-1.0em}
    \includegraphics[width=\linewidth]{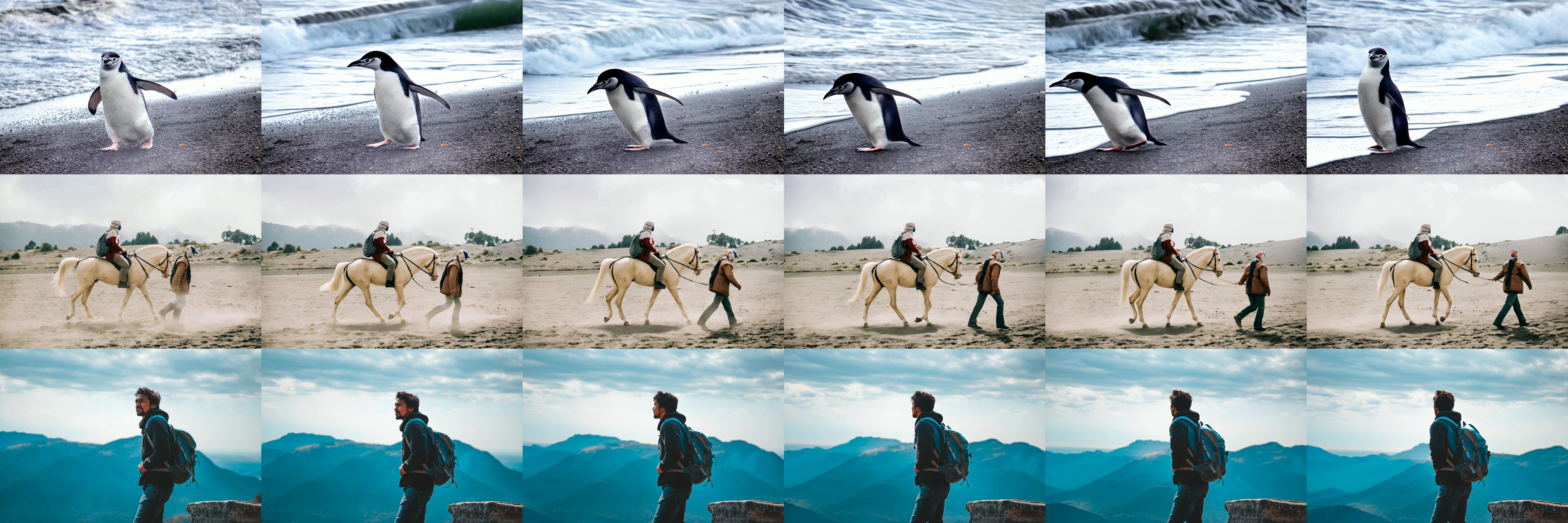}\vspace{-0.5em}
    \caption{Videos generated via FSVideo's Image-to-Video framework while being $\mathbf{42.3\times}$ faster than Wan2.1-I2V-14B-720P. For every row, first frame is the input image from VBench~\citep{huang2024vbench} testset, and the following frames are generated. Zoom in to see details.} \vspace{-1.2em}
    \label{fig:banner}
\end{figure}
\section{Introduction}
\input{Sections/1_introduction}
\section{Method}
\subsection{Framework Overview}
\label{subsec:fsVideo_overview}
\input{Sections/2_1_overall_design}
\subsection{Video Autoencoder}
\label{subsec:AE}
\input{Sections/2_2_autoencoder}
\subsection{Video Diffusion Transformer}
\label{subsec:dit}
\input{Sections/2_3_video_dit}

\subsection{Video Upsampler}
\label{subsec:upsampler}
\input{Sections/2_4_video_upsampler}

\section{Experiments}
\label{sec:exp}
\subsection{Implementation Details}
\label{subsec:implementation_details}
\input{Sections/3_1_implementation_details}
\subsection{Evaluation}
\input{Sections/3_2_evaluation}
\section{Conclusion}
\label{sec:conclusion}
\input{Sections/4_conclusion}
\section{Contributors}
\label{sec:contributors}
Contributor names are alphabetically listed by last name and then first name. Names with an asterisk (*) are people who have left the company.
\paragraph{Core contributors:} Xinwei Huang, Minxuan Lin, Yaojie Shen, Xiao Yang*, Yuxin Zhang
\paragraph{Contributors:}
Qingyu Chen*, Zhiyuan Fang, Haibin Huang*, Tong Jin, Bo Liu, Celong Liu*, Chongyang Ma, Xing Mei*, Xiaohui Shen, Fuwen Tan, Angtian Wang, Yiding Yang, Jiamin Yuan, Lingxi Zhang

\clearpage

\bibliographystyle{plainnat}
\bibliography{main}

\clearpage



\end{document}

%% file: math_commands.tex
\usepackage{amsmath,amsfonts,bm}

\newcommand{\batch}{{\textit{B}}}
\newcommand{\frames}{{\textit{F}}}
\newcommand{\height}{{\textit{H}}}
\newcommand{\width}{{\textit{W}}}
\newcommand{\tokens}{{\textit{T}}}
\newcommand{\dims}{{\textit{D}}}
\newcommand{\layers}{{\textit{L}}}
\newcommand{\hiddenstates}{{\mathbf{X}}}
\newcommand{\keys}{{\mathbf{K}}}
\newcommand{\values}{{\mathbf{V}}}
\newcommand{\queries}{{\mathbf{Q}}}
\newcommand{\latent}{{\mathbf{\textit{z}}}}
\newcommand{\velocity}{{\textit{v}}}

\def\eqref#1{equation~\ref{#1}}

\def\1{\bm{1}}

\DeclareMathAlphabet{\mathsfit}{\encodingdefault}{\sfdefault}{m}{sl}
\SetMathAlphabet{\mathsfit}{bold}{\encodingdefault}{\sfdefault}{bx}{n}



%% file: Sections/1_introduction.tex
\begin{figure}[t]
    \centering\vspace{-1.0em}
    \includegraphics[width=\linewidth]{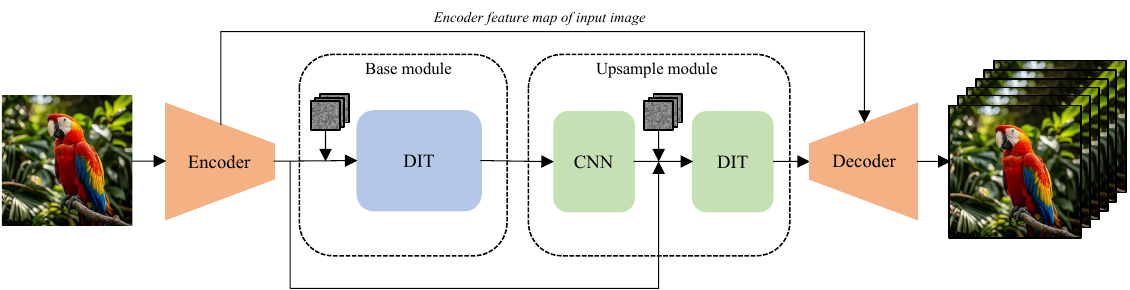}\vspace{-0.5em}
    \caption{Overall framework of FSVideo image-to-video pipeline. The input image is sent to the encoder to get its VAE latent code, which is then used as the condition of the base module DIT for the first diffusion process. Then, the diffused latent is sent to the upsample module, where it first passes a latent convolution neural net (CNN) upscaler, and is then combined with the input image's latent as the condition to the upsampler DIT for another diffusion process. Finally, the upsampled latent is sent to the decoder to generate output video. The decoder contains attention layers conditioned on input image's encoder feature map to enhance video quality (see Section~\ref{subsubsec:decoder_improvement}).} \vspace{-1.2em}
    \label{fig:overall_framework}
\end{figure}
Video generation models have become a major focus in both academia and industry, due to their strong generation ability and wide range of application potential. Although recent models such as Sora 2~\citep{openai2025sora2}, Veo 3~\citep{google2025veo3}, Kling~\citep{kuaishou2025Kling}, Wan~\citep{wan2025wanopenadvancedlargescale} and SeedDance~\citep{gao2025seedance10exploringboundaries} have shown impressive generation capability, the high inference cost of these large video models often results in long waiting times and high GPU cost, making scaling to a broader audience a challenge.

Methods that increase video model generation speed, especially for models based on diffusion and flow-matching formulation, have been a very popular research topic~\citep{shen2025efficient}. Many works have focused on training-free acceleration methods, such as efficient solvers~\citep{dpmsolver,unipc}, cache-based methods~\citep{teacache,fastercache}, low-resolution sampling~\citep{tian2025trainingfreediffusionaccelerationbottleneck} and sparse attention operation~\citep{xi2025sparse,chen2025sparsevditunleashingpowersparse}. However, the speed-up achieved via these training-free methods is often limited and may result in generation quality degradation. 

In a different direction, training-based acceleration methods have also been extensively explored. Some methods are proposed to reduce the model size~\citep{wu2025taming}, or replace attention operation with lightweight alternatives~\citep{becker2025edit, chen2025sana,huang2025m4v}, but these methods usually suffer from loss of generation quality due to reduced model capacity. Step-distillation-based methods can get greater speed-up while keeping the quality of the generation. The majority of these methods focus on the distillation of inference steps~\citep{zhai2024motion,zhang2024sf,lin2025diffusion,mao2025osv,yin2025slow}, reducing the number of inference steps to a single digit number or even 1, offering more than an order of magnitude of speed-ups. However, in industry usage, it is more common to use a inference step that is larger than 1 (e.g., 4 to 8 step), since the generation quality drastically degrades when reaching very low inference steps due to the high estimation error of the latent ODE/SDE path. Thus, when the undistilled model uses a reasonable amount of inference steps~\footnote{For example, Wan 2.1 by default uses UniPC solver~\citep{unipc} with 50 inference steps in its open source repository \url{https://github.com/Wan-Video/Wan2.1}} and an efficient solver, the speed-up achieved from step distillation without sacrificing generation quality can be limited. This, combined with the high amount of computation per model forward, makes video diffusion model computation intensive.

In this paper, we propose FSVideo, a video diffusion transformer model designed for fast image-to-video generation. The key idea of FSVideo is to reduce the amount of compute per model forward. To achieve this, we propose a video autoencoder (AE) with a spatial-temporal downsampling of $64\times64\times4$, thus reducing the overall video generation time. Some previous work~\citep{hacohen2024ltx,peng2025open} uses video autoencoder with up to $32\times$ spatial compression ratio, but they may suffer from poor reconstruction and generation quality. There is a concurrent work~\citep{chen2025dc} that proposes a deep compression video autoencoder with up to $64\times64$ spatial compression ratio. However, generative quality evaluation is done mainly using the autoencoder with $32\times32$ spatial compression ratio. In addition, \citep{chen2025dc} mainly focuses on fine-tuning a pretrained video diffusion transformer (DIT) to the new VAE latent space, while in this work we explore training a DIT from scratch in a highly compressed latent space. Of course, our proposed new techniques can also be applied by fine-tuning an existing video DIT.

Our main contributions in FSVideo are:
\begin{itemize}
\item \textbf{Video autoencoder with highly compressed latent space}: we introduce \textbf{FSAE}, a new asymmetric video autoencoder with $64\times64\times4$ spatial-temporal compression with 128 latent channels, resulting in a total of $384\times$ information reduction. This autoencoder achieves competitive reconstruction performance while having strong generation capability.
\item \textbf{Improved diffusion transformer with layer memory}: we propose a new diffusion transformer (DIT) architecture with a \textbf{layer memory} mechanic to provide more freedom to the DIT information flow during the diffusion process. This results in better utilization of the DIT model capacity and better generation performance, with negligible overhead.
\item \textbf{Latent video upsampler for increased video fidelity}: we propose a upsampler module, consists of a Convolutional latent upsampler and a DIT refiner, to upscale base DIT's latent output, drastically increasing video fidelity while ensuring minimum inference time increase via a lightweight refiner DIT step distillation.
\item Benefiting from our design, FSVideo is able to generate videos with competitive quality while being an order of magnitude faster compared to other open-source models of similar amount of parameters (e.g. $\mathbf{42.3\times}$ faster than Wan2.1-14B).
\end{itemize}
The reminder of the report are organized as follows. Section~\ref{subsec:fsVideo_overview} discusses the overall FSVideo framework, then introduces the new Video AE (Section~\ref{subsec:AE}), base DIT (Section~\ref{subsec:dit}) and upsampler (Section~\ref{subsec:upsampler}) in detail. Section~\ref{sec:exp} presents implementation details of the training framework, as well as the experimental results of both VAE reconstruction and video generation. Section~\ref{sec:conclusion} discusses the potential extension and applications of our method.

%% file: Sections/2_1_overall_design.tex
The general framework of FSVideo is shown in Figure~\ref{fig:overall_framework}. Note that video generation model can have various input types, such as text-only (text-to-video), text with first frame (image-to-video), and text with first and last frame (first-last-iamge-to-video), etc. We formulate FSVideo as an image-to-video model for the following reasons:
\begin{enumerate}
    \item \textbf{Ease of training}: We aim to train diffusion transformers in a heavily compressed latent space with a high number of latent channels (128 in our case). Diffusion training in such a latent space has been shown to be a hard task~\citep{chen2025deep, chen2025dc1.5}. By constraining the training task to be only on image-to-video task, which receives ample information of the video appearance via the input image being the video's first frame, we let the diffusion training focus more on modeling video movement, reducing the training difficulty and advocating better model performance. Moreover, there are tricks specifically suitable for image-to-video that further boost generation quality, such as in Section~\ref{subsubsec:decoder_improvement} and Section~\ref{subsec:high_res_refiner}.
    \item \textbf{Resource constraints}: The development of FSVideo is challenged by limited resources in GPU compute, training data, and labeling workforce. Thus, it is hard to gather sufficient numbers of high-quality video data with good aesthetics for text-to-video. Image-to-video is more data-friendly, and video aesthetics can be guaranteed by using a high-quality text-to-image model to generate the first frame. 
    \item \textbf{Application consideration}: Many applications, such as photo reenactment and visual effects, often rely on users providing an image in the real world as the first frame of the video, which fits naturally in the image-to-video setting. For cases where input image is not required, we can always fall back to the text-to-image-to-video pipeline stated above.
\end{enumerate}

%% file: Sections/2_2_autoencoder.tex
Here we introduce our new FSAE model. We first present the model architecture and training design, then discuss methods to balance generation and reconstruction quality for the autoencoder's latent space, and finally talk about various ways to improve the decoder quality, where we introduce two variants of FSAE: FSAE-Standard, which leans on reconstruction quality, and its lightweight variant FSAE-Lite, which mildly sacrifices quality for great speed and memory-consumption improvement.

\begin{figure}[t]
    \centering\vspace{-1.0em}
    \includegraphics[width=\linewidth]{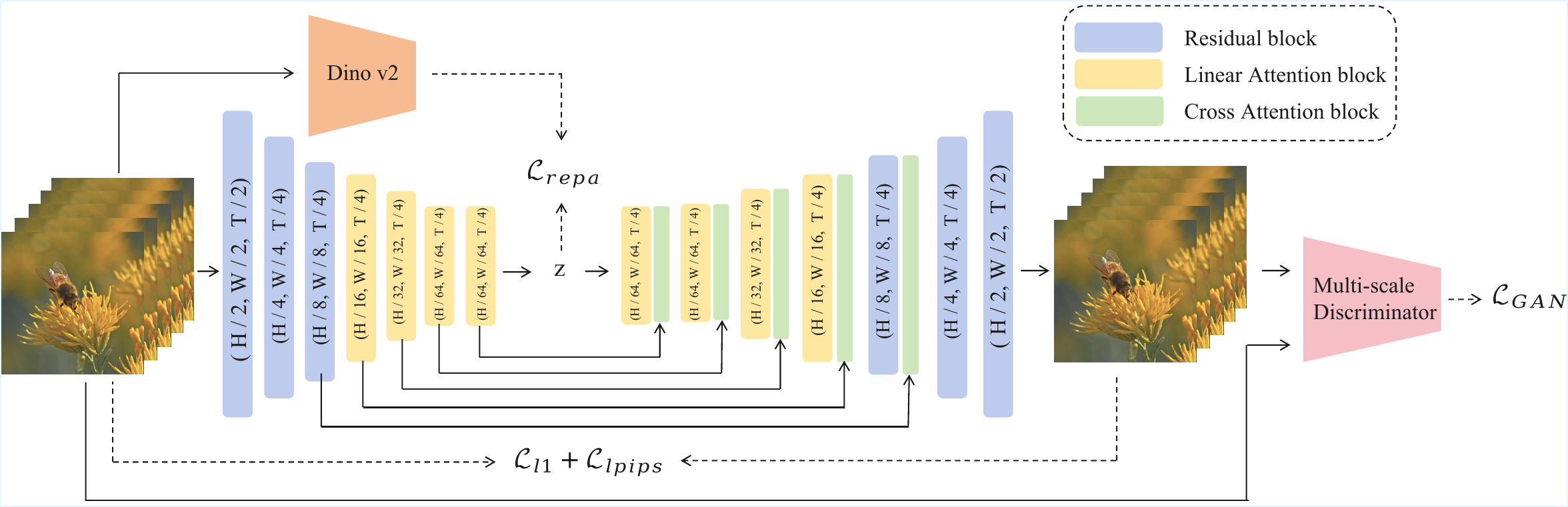}\vspace{-0.5em}
    \caption{Overall framework of FSAE.} \vspace{-1.2em}
    \label{fig:vae_framework}
\end{figure}
\subsubsection{Overall Design And Training Strategy}
\label{subsubsec:overall_vae_desigin_and_training}
First we define the notation of the total compression rate of an autoencoder. Given a video $V \in \mathbb{R}^{3 \times T \times H \times W}$, the encoder $E$ encodes it into a latent $Z \in \mathbb{R}^{c \times t \times h \times w}$, and decodes the latent via the decoder $D$ to recover the original video. The spatial compression $f_h$ and $f_w$ is calculated by $h/H$ and $w/W$, and the temporal compression is $f_t=(t-r_t)/(T-r_t)$ where $r_t$ is set to 1 if using causal convolution else 0. The total compression ratio is formulated as follows:
\begin{equation}
    \textit{Total\_Compression} = f_h\cdot f_w\cdot f_t \cdot\frac{3}{c}
\end{equation}

\paragraph{Model structure.} Figure~\ref{fig:vae_framework} shows our FSAE framework. Note that, different from LTX-Video, we do not apply patchify to the input video as a preprocessing step, and all downsample/upsample operations are done inside the autoencoder. This shows~\citep{tian2024reducio} better reconstruction quality and less artifacts. Our video autoencoder follows DC-AE~\citep{chen2025deep}, which is a deep-compressed autoencoder for image generation. We start from the \textsl{dc-ae-f32c32-sana-1.0} version of DC-AE, where both the encoder and the decoder consist of 3 convolution blocks and 3 transformer blocks, achieving a $32\times32$ spatial compression ratio. We first add an additional set of transformer blocks to both the encoder and the decoder, bringing the spatial compression ratio to $64\times64$, and increase the latent channel to 128. Then we expand the 2D convolution kernels of the autoencoder to causal 3D convolutions, which enables long video generation and joint training between video and image~\citep{wan2025wanopenadvancedlargescale,kong2024hunyuanvideo}. We then change the downsample/upsample operations~\cite{shi2016real} to include temporal dimension, enabling time-spatial-to-channel and channel-to-time-spatial transition. Finally, we change the downsample/upsample operation of the outer two blocks of encoder/decoder to include temporal dimension reduction, making the autoencoder performing $4\times$ temporal compression, achieving our compression target of $64\times64\times4$. We also make some minor adjustments to the model architecture, such as making the attention layers in the autoencoder operate only on height and width dimension, as well as changing all norms to pixel norm. These changes ensure that the latent computation is not sensitive to time and spatial dimension changes and allow us to apply various memory reduction inference techniques, such as temporal splitting and 3D tiling.

\paragraph{Training Strategy.} We apply a multi-stage training strategy to the autoencoder. During the first stage, we train on $256\times256$ resolution\footnote{Here $256\times256$ indicates the average resolution, with varying aspect ratios. Training data is grouped via an aspect ratio bucketing strategy, with image/videos of similar aspect ratios cropped and resized to the same resolution bucket. The same applies in later sections.}, using both images and 17-frame videos. We use both $L1$ and LPIPS loss~\citep{johnson2016perceptual} in this stage. In the second stage, we add GAN loss to improve video quality. We use a 3D multiscale discriminator and use non-saturating logistic loss~\citep{goodfellow2014generative} with R1 regularization~\citep{mescheder2018training}. In this stage, we also increase the image/video resolution to $512\times512$, and increase the video duration to 61 frames. In the final stage, we further scale the spatial and temporal resolutions up to 121-frame $1024\times1024$ videos. Each stage is trained for approximately $200,000$ iterations. The final training loss $\mathcal{L}_{ae}$ is written as $\mathcal{L}_{ae} = \mathcal{L}_1 + 0.1\times\mathcal{L}_{lpips} + 0.1\times\mathcal{L}_{GAN}$.

Training autoencoder on high-resolution videos can be resource-challenging, which is especially true for our case, since we do not apply patchify to the input video, drastically increasing the feature map size. Thus, in stage 3 of our autoencoder training, we apply several tactics to reduce memory consumption. First, we apply a mixed-resolution training strategy by training the autoencoder with both low-resolution, long-duration videos (e.g. $256\times256\times171$) and high-resolution, short-duration ones (e.g. $704\times704\times9$). This strategy leads to a better generalization to high-resolution and long-duration reconstruction~\citep{seawead2025seaweed7bcosteffectivetrainingvideo}. Second, when training on very high spatial resolutions (e.g. $1024\times1024$), we randomly extract a small 3D patch from the full feature map from the third-to-last block of the decoder, and only forward this small patch to the remaining 3 decoder blocks. We then only use the output of this small patch for training loss computation and gradient back-propagation.\footnote{Due to fixed padding of convolution layers in the spatial dimension, we will get different results between patch-based generation and full frame generation. Thus we only calculate loss using the patch area with correct boundary conditions.} This is similar to applying a 3D binary window to the output video, and only calculate loss inside that window, but our approach can greatly reduce training memory while achieve the same effect. Finally, we apply temporal slicing for LPIPS calculation to reduce peak memory usage. Together, these techniques help drastically reduce the memory usage of training during the final stage.

\input{tables/instrinsic_dimension}
\subsubsection{Latent Space Sematic Alignment}
Since our final goal is to use the video autoencoder for DIT training, it is important to ensure that the autoencoder has good generation capability. Several methods~\citep{yao2025reconstruction,kouzelis2025eqvae,skorokhodov2025improving} have been proposed to improve the semantic alignment of the latent space for the image autoencoder, but no discussions have been done in the video space yet. Here we propose \textbf{Video} \textbf{V}ision \textbf{F}oundation model alignment Loss (\textbf{Video VF Loss}), by extending VA-VAE~\citep{yao2025reconstruction}'s paradigm to videos, as described below.

Given a video $V$, we obtain the latent $Z \in \mathbb{R}^{c \times t \times h \times w}$ through the encoder, and use Dinov2~\citep{oquab2023dinov2} to extract frame-by-frame feature of the video to obtain $F \in \mathbb{R}^{c' \times t' \times h' \times w'}$, and we want to align $Z$ with $F$ to improve the generation ability of the video autoencoder. A natural question is how to align two feature maps with different channel, spatial, and temporal dimensions. To solve this, we propose a series of operations to match the feature dimensions. First, we map the channel dimension of $Z$ to $F$ via a learnable linear layer $\textbf{W} \in \mathbb{R}^{c' \times c}$. Then, we interpolate $Z$ on the spatial dimension $h$ and $w$ to match $h'$ and $w'$. Finally, since $Z$ has a smaller $t$ dimension due to temporal compression, we apply average pooling of kernel size 4 to $F$ in $t'$ dimension\footnote{The first video frame's feature in $F$ does not perform this pooling operation due to the video AE's causal nature.}. After these steps,we obtain a aligned feature pair $(z, f)$ with the same shape $c' \times t \times h' \times w'$.

After the features are aligned, we then expand the feature matching loss of \citep{yao2025reconstruction} to video domain as Video Marginal Cosine Similarity Loss $\mathcal{L}_{v-mcos}$
\begin{equation}
\mathcal{L}_{v-mcos}=\frac{1}{t \times h' \times w'} \sum_{i=1}^{t} \sum_{j=1}^{h'} \sum_{k=1}^{w'} ReLU\left(1-m_{1}-\frac{z_{i j k} \cdot f_{i j k}}{\left\| z_{i j k}\right\| \left\| f_{i j k}\right\| }\right).
\end{equation}
and Video Marginal Distance Matrix Similarity Loss $\mathcal{L}_{v-mdms}$
\begin{equation}
\mathcal{L}_{v-mdms}=\frac{1}{(t\times h'\times w')^{2}} \sum _{p,q}^{t\times h'\times w'}ReLU\left(\left|\frac{z_{p} \cdot z_{q}}{\| z_{p}\| \| z_{q}\| }-\frac{f_{p} \cdot f_{q}}{\| f_{p}\| \| f_{q}\| }\right|-m_{2}\right).
\end{equation}

The autoencoder training loss with the Video VF Loss is then written as
$\mathcal{L}_{ae}$ and $\mathcal{L}_{v-vf}$:
\begin{equation}
\mathcal{L}_{total}=\mathcal{L}_{ae}+\mathcal{L}_{v-vf}=\mathcal{L}_{ae}+\alpha \left(\mathcal{L}_{v-mcos}+\mathcal{L}_{v-mdms}\right).
\end{equation}

with $m_1=0.5, m_2=0.25, \alpha=0.5$. We integrate this loss by finetuning the autoencoder with $\mathcal{L}_{total}$.

To check our method's impact on the complexity of the latent manifold, we calculate the intrinsic dimension of the latent space. Intrinsic dimension evaluates the minimum number of variables needed to represent a distribution~\citep{intrinsic_dimension_signal_collection}. Hence, a lower intrinsic dimension indicates lower latent space complexity. We use Gride~\citep{denti2022generalized} algorithm with up to 64 neighbors from the Dadapy~\citep{glielmo2022dadapy} package for intrinsic dimension calculation. For comparison, we also implement other regularization techniques proposed in \citep{kouzelis2025eqvae,skorokhodov2025improving} and calculate their intrinsic dimension. The result is shown in Table~\ref{tab:instrinsic_dimension}. It can be seen that our proposed Video VF loss achieves the overall lowest intrinsic dimension, while being drastically better than the autoencoder without any regularization. This shows that our Video VF loss can greatly reduce the latent space complexity, therefore improving the generation ability of the autoencoder.
\subsubsection{Decoder Improvement}
\label{subsubsec:decoder_improvement}
Due to the high compression ratio of our video autoencoder, we still see two problems after applying the training discussed above: artifacts in the reconstructed videos and high computation requirement during the decoding stage. Therefore, we extend our autoencoder to an asymmetric structure, where we freeze the encoder, make architecture changes to the decoder, and only finetune the decoder. Below we discuss the steps we take for video quality enhancement and decoding performance optimization on the decoder during this stage.

\paragraph{Video Quality Enhancement.}
To improve video quality and reduce artifacts, we implement the following architecture changes to the video decoder:

\begin{enumerate}
    \item Change the convolution layers in the decoder to non-causal. We find that causal convolution introduces noticeable frame flickering, due to its unidirectional temporal dependency, especially in the decoder part. Switching these layers to non-causal convolutions solves this problem.
    \item Inject first-frame features from encoder to decoder. Since we only focus on image-to-video generation, we can utilize the input image as an additional condition to the decoder to further increase the reconstruction quality. Past work like Reducio-VAE~\citep{tian2024reducio} explored similar ideas, but Reducio-VAE uses an extra 2D VAE to encode mid-frame feature, while we directly reuse the FSAE encoder to encode first-frame feature, which removes the need for an additional encoder. This is enabled due to the causal property of the encoder, which handles first frame independently from following frames. As shown in Figure~\ref{fig:vae_framework}, we extract first-frame features from the last 5 blocks of the encoder, and insert the feature into the corresponding decoder blocks via cross-attention.
    \item Noise-injection~\citep{hacohen2024ltx} to each convolution block. This helps with high-frequency detail generation. We use standard Gaussian noise with a fixed weight of 0.05.
\end{enumerate}
We apply these optimizations as an additional finetuning step to the FSAE model, and achieve our official video autoencoder, which we call \textbf{FSAE-Standard}. This version has the best reconstruction quality, but may suffer from a long generation time and high memory consumption. Therefore, we design another version of FSAE with less compute at the cost of a slightly degraded performance, which we discuss below.

\paragraph{Performance Optimization.}
Starting from FSAE-Standard, we inspect the memory consumption, and find the main performance bottlenecks to be the last two blocks, which are closer to the RGB space and therefore having the largest feature map. Thus, we reduce the number of channels in these blocks, decreasing the maximum memory usage.

We also reconsider the choice of convolution in the decoder. As stated in \textsl{Video Quality Enhancement} section, causal convolution may introduce temporal flickering, while noncausal convolution has high memory cost since we can no longer perform temporal slicing like we can with causal ones. Therefore, we take the middle ground and use group-causal convolution~\citep{wu2025improved} in the decoder. Depending on different temporal compression ratio in each decoder block, we use group size of 1, 2 and 4 for the group-causal layers, respectively. Besides, different from \citep{wu2025improved}, we use replication padding after each frame group, making finetuning much easier.

We finetune FSAE-Standard with these changes to get the \textbf{FSAE-Lite} version of the autoencoder, which has 1.75-2$\times$ less memory consumption and less inference time.

\input{tables/vae_quality_comparsion}
\begin{figure}[h!]
    \centering\vspace{-1.0em}
    \includegraphics[width=\linewidth]{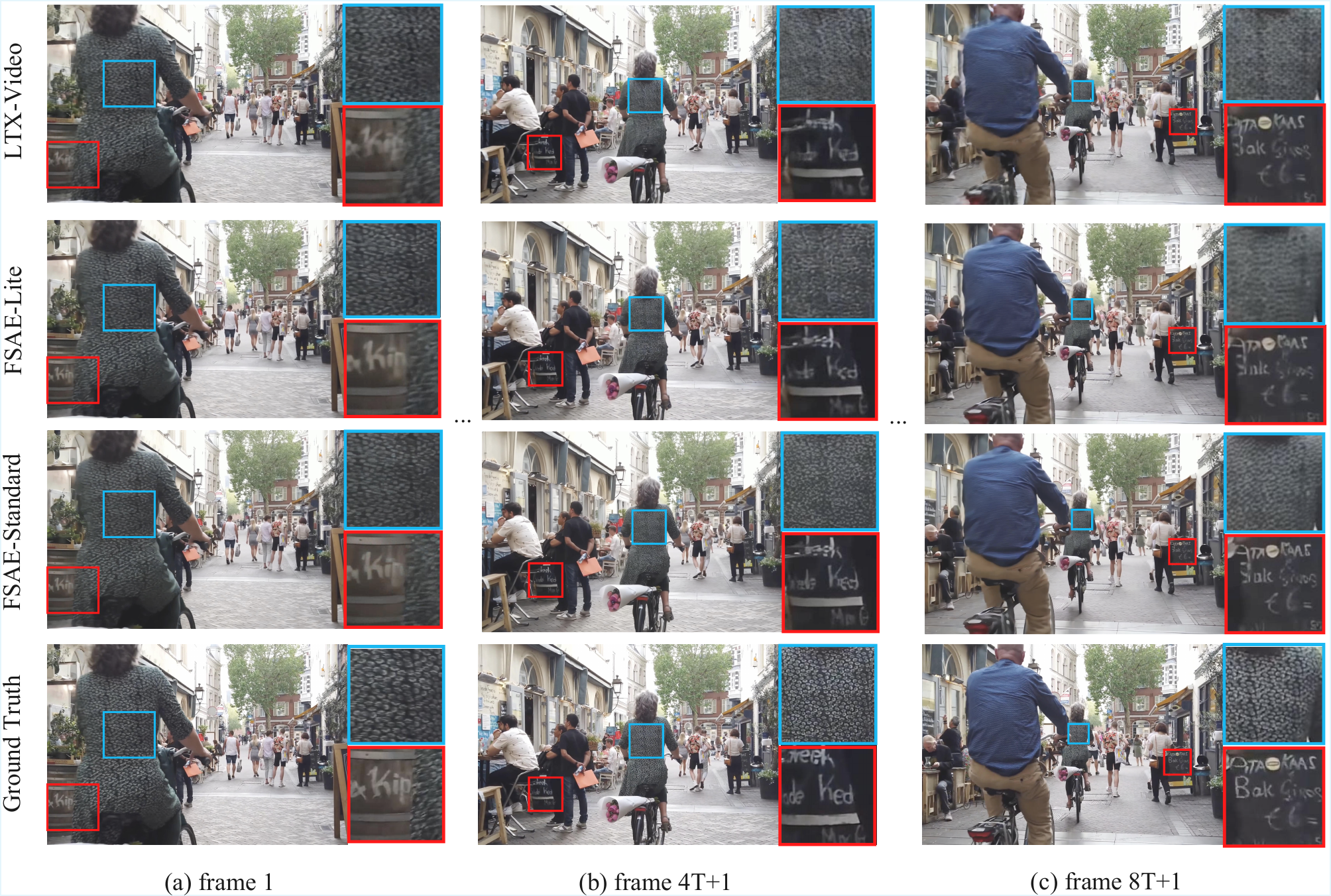}\vspace{-0.5em}
    \caption{Video reconstruction comparison between LTX-Video's autoencoder and FSAE. The first three rows represent the reconstruction results of different AE, and the last row is the ground truth. The blue box tracks the clothing textures' temporal consistency, where LTX-Video exhibits inconsistent inter-frame flickering, and FSAE consistently maintains the dotted texture. The red box compares the reconstruction quality to video details: FSAE-Lite and LTX-Video achieve comparable reconstruction results, whereas FSAE-Standard outperforms LTX-Video.} \vspace{-1.2em}
    \label{fig:vae_visual_comparison}
\end{figure}
\paragraph{Qualitative and Quantitative Measurement} We show the comparison between FSAE and other state-of-the-art video autoencoder methods in Table~\ref{tab:vae_quality_comparisons}. We evaluate all methods using $256\times256\times17$ videos using SSIM~\citep{bovik2004image}, PSNR, LPIPS~\citep{johnson2016perceptual} and FVD~\cite{unterthiner2018towards}. The evaluation set includes 1000 unseen videos from Inter-4K~\citep{stergiou2022adapool} and 1000 unseen validation set videos from WebVid-10M~\citep{bain2021frozen}. As shown in the table, FSAE achieves better reconstruction accuracy compared to other autoencoders with high compression rate, such as LTX-Video and VidTok~\citep{tang2024vidtok}, and can even outperform some AEs with low compression rate, such as Cosmos-CV~\citep{agarwal2025cosmos}. We also visualize the video reconstruction result in Figure~\ref{fig:vae_visual_comparison}, showing FSAE's competitive visual reconstruction ability.

%% file: tables/instrinsic_dimension.tex
\begin{table}[t]

\centering
\caption{FSAE's Intrinsic dimension comparisons with different regularization methods. Lower number $=$ lower latent space complexity.}
\resizebox{\textwidth}{!}{
    \begin{tabular}{l|c|c}
    \toprule
    \multirow{2}{*}{\textbf{Method}} & \multicolumn{2}{c}{\textbf{Intrinsic Dimension using Gride~\citep{denti2022generalized} with 2/4/8/16/32/64 nearest neighbors}} \\
    & Video$\downarrow$ & Image$\downarrow$ \\
    \midrule
    No regularization & [87.83, 182.16, 11.22, 16.94, 24.32, 29.55] & [33.15, 62.24, 102.51, 12.07, 15.45, 21.13] \\
    Downscale regularization & [30.5, 37.48, 31.99, 25.86, 25.34, 24.05] & [25.73, 38.20, 41.52, 28.75, 26.66, 26.59] \\
    Upscale regularization & [46.63, 57.16, 28.66, 28.11, 30.8, 29.63] & [34.92, 44.30, 44.17, 26.39, 29.02, 27.97] \\
    \rowcolor{blue!10} Video VF & [24.44, 29.60, 26.93, 21.11, 22.15, 22.04] & [21.46, 24.52, 24.91, 18.07, 20.48, 20.13] \\
    \bottomrule
    \end{tabular}
}
\label{tab:instrinsic_dimension}
\end{table}

%% file: tables/vae_quality_comparsion.tex
\begin{table}[t!]

\centering
\caption{Quantitative comparisons with other SOTA methods and our FS Video Autoencoder.}
\resizebox{\textwidth}{!}{
    \begin{tabular}{l|c|c|cccc|cccc}
    \toprule
    \multirow{2}{*}{\textbf{Method}} & \multirow{2}{*}{\textbf{\makecell{Downsample\\Factor}}} & \multirow{2}{*}{\textbf{\makecell{Total\\Compression}}} & \multicolumn{4}{c}{\textbf{Inter-4K}} & \multicolumn{4}{c}{\textbf{WebVid-10M}} \\
    & & & SSIM$\uparrow$ & PSNR$\uparrow$ & LPIPS$\downarrow$ & FVD$\downarrow$ & SSIM$\uparrow$ & PSNR$\uparrow$ & LPIPS$\downarrow$ & FVD$\downarrow$ \\
    \midrule
    Hunyuan VAE~\citep{kong2024hunyuanvideo} & 8 $\times$ 8 $\times$ 4 & 1:48 & 0.891 & 32.56 & 0.047 & 73.50 & 0.923 & 33.99 & 0.031 & 63.72 \\
    Wan-2.1 VAE~\citep{wan2025wanopenadvancedlargescale} & 8 $\times$ 8 $\times$ 4 & 1:48 & 0.880 & 31.73 & 0.049 & 79.29 & 0.913 & 33.12 & 0.032 & 71.83 \\
    CogVideoX-1.5 VAE~\citep{yang2025cogvideox} & 8 $\times$ 8 $\times$ 4 & 1:48 & 0.880 & 31.44 & 0.072 & 109.11 & 0.920 & 33.53 & 0.043 & 80.98 \\
    Step-Video VAE~\citep{ma2025stepvideot2vtechnicalreportpractice} &  16 $\times$ 16 $\times$ 8 & 1:96 & 0.847 & 30.20 & 0.082 & 125.45 & 0.894 & 31.86 & 0.050 & 96.75 \\
    VidTok~\citep{tang2024vidtok} & 8 $\times$ 8 $\times$ 4 & 1:96 & 0.835 & 29.98 & 0.081 & 159.88 & 0.887 & 31.79 & 0.048 & 124.46 \\
    \midrule
    Cosmos-CV~\citep{agarwal2025cosmos} & 8 $\times$ 8 $\times$ 8 & 1:96 & 0.805 & 29.16 & 0.184 & 342.99 & 0.864 & 30.76 & 0.113 & 211.33 \\
    LTX-Video~\citep{hacohen2024ltx} & 32 $\times$ 32 $\times$ 8 & 1:192 & 0.787 & 28.40 & 0.153 & 370.86 & 0.868 & 30.89 & 0.077 & 232.02 \\
    VidTok~\citep{tang2024vidtok} & 8 $\times$ 8 $\times$ 4 & 1:192 & 0.783 & 28.20 & 0.119 & 266.07 & 0.843 & 29.85 & 0.077 & 217.77 \\
    Cosmos-CV~\citep{agarwal2025cosmos} & 16 $\times$ 16 $\times$ 8 & 1:384 & 0.724 & 26.72 & 0.271 & 704.08 & 0.786 & 28.09 & 0.180 & 518.86 \\
    VidTok~\citep{tang2024vidtok} & 16 $\times$ 16 $\times$ 4 & 1:768 & 0.645 & 24.19 & 0.211 & 680.41 & 0.713 & 25.56 & 0.138 & 625.32 \\
    \midrule
    \midrule
    \rowcolor{blue!10} FSAE-Standard & 64 $\times$ 64 $\times$ 4 & 1:384 & 0.806 & 28.96 & 0.107 & 256.62 & 0.872 & 30.91 & 0.058 & 203.19 \\
    \rowcolor{blue!10} FSAE-Lite & 64 $\times$ 64 $\times$ 4 & 1:384 & 0.788 & 28.48 & 0.151 & 342.66 & 0.861 & 30.42 & 0.075 & 240.03 \\
    \bottomrule
    \end{tabular}
}
\label{tab:vae_quality_comparisons}
\end{table}

%% file: Sections/2_3_video_dit.tex
\begin{wrapfigure}{R}{0.41\textwidth}
\vspace{-1.0em}
\includegraphics[width=0.99\linewidth]{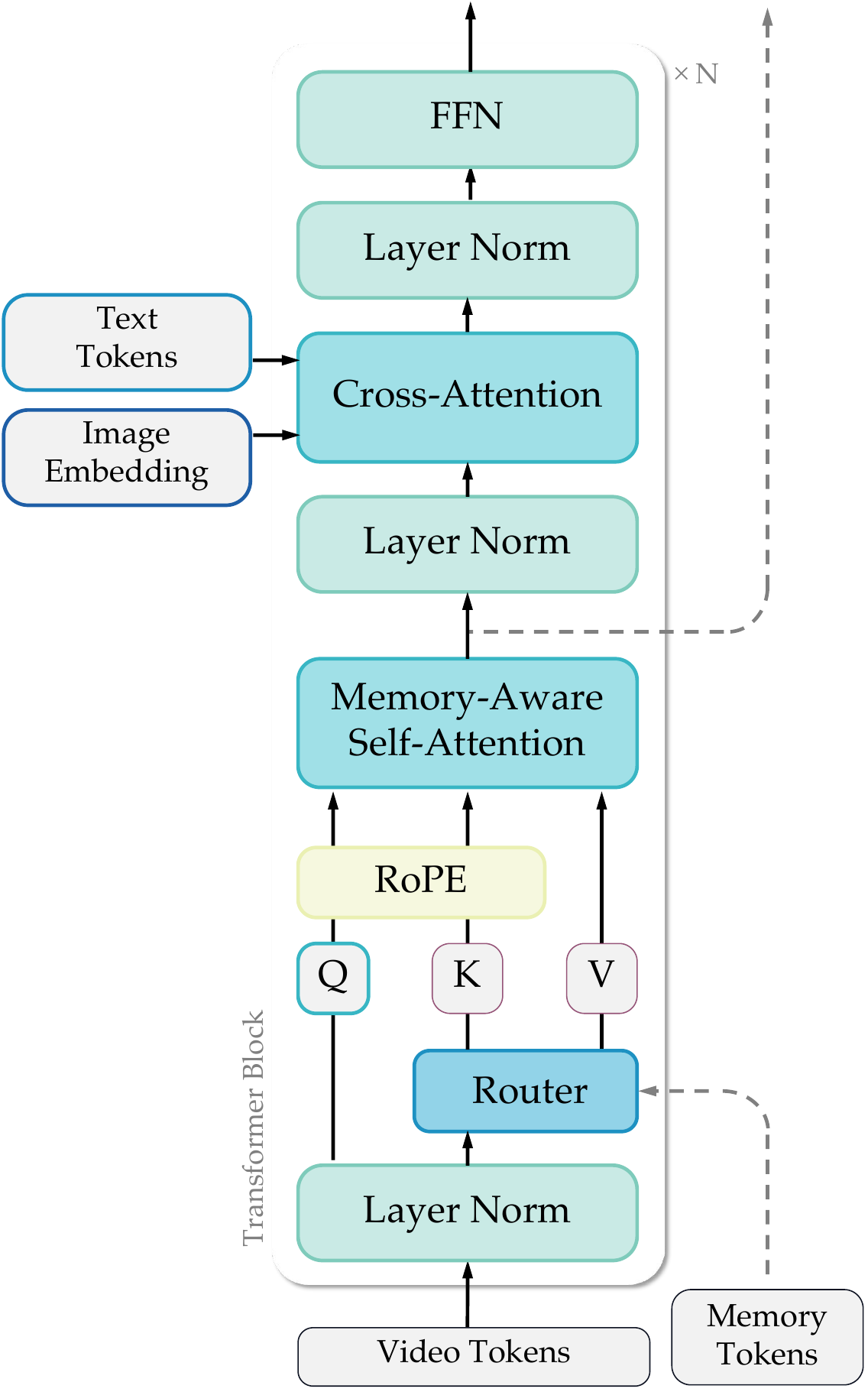} 
\caption{Transformer layer of FSVideo.}
\vspace{-1.5em}
\label{fig:transformer}
\end{wrapfigure}
Since we train our DIT in a highly compressed latent space, we want to utilize a mature DIT architecture to ensure stable training. 
At the same time, we also want to explore ways to better utilize DIT's representation capability. 
Based on this mindset, we choose Wan2.1-14B-I2V~\citep{wan2025wanopenadvancedlargescale}'s DIT structure as the baseline and add modifications on it.
Here we very briefly discuss Wan2.1-14B-I2V's DIT
structure and the changes we make to it. The DIT consists of a patchify module via 3D convolution, a series of transformer layers, and an unpatchify module. In the patchify module, unlike WAN, we use a kernel size of $(1, 1, 1)$ since we have already performed heavy compression in the autoencoder. This results in a sequence of latent tokens with shape $\left(\batch, \tokens, \dims\right)$,
where $\batch$ denotes the batch size, $\dims$ the latent embedding dimension and
$\tokens = (1 + \frames / 4) \times \height / 64 \times \width / 64$ the sequence length. The improved transformer layer with layer memory is shown in Figure ~\ref{fig:transformer}, where the transformer layer contains both self-attention and cross-attention modules conditioned on both UmT5~\citep{umt5}'s text embedding  and CLIP~\citep{clip} embedding of the input image. To improve inter-layer information flow and strengthen DIT's representation capacity, we introduce a new \emph{Layer Memory Self-Attention} mechanism (see Section~\ref{subsec:layer_memory}) that allows each layer to adaptively attend to partial attention features from all preceding layers.
This design facilitates hierarchical information reuse and enhances temporal coherence across depth.

\subsubsection{Layer Memory Self-Attention Mechanism}
\label{subsec:layer_memory}
In language modeling research, various methods~\citep{fang2023cross,zhu2024hyper,lime} have been proposed to combat representation collapse, a problem in transformers where features of adjacent layers are excessively similar, limiting the transformer's representation capacity, especially for very deep networks~\citep{representation_collapse}. However, similar attempts are rarely explored in image and video generation tasks, with U-DIT~\citep{tian2024u} being the closest work. In U-DIT, the DIT is formulated into a U-net style structure, with encoding layers downsampling the tokens and decoding layers upsampling them, and skip connections are added symmetrically at each downsampling/upsampling transition. We want to explore a method that is applicable to general DIT structure, thus we propose to integrate the \emph{Layer Memory} mechanism into DITs.
Unlike the standard design where each layer attends only to the preceding hidden states, Layer Memory enables every self-attention block to access a learnable mixture of all prior layer representations, thereby forming a differentiable memory across DIT depth.

In a conventional $\layers$-layer decoder-only DIT, the $l$-th self-attention layer ($1 \leq l \leq \layers$) updates its representation as:
$\hiddenstates_{l} = \mathrm{SelfAttention}(\hiddenstates_{l-1})$,
where $\hiddenstates_{l-1} \in \mathbb{R}^{\batch \times \tokens \times \dims}$.
The multihead attention mechanism projects $\hiddenstates_{l-1}$ into queries, keys, and values:
$\queries, \keys, \values \in \mathbb{R}^{n \times d}$, with $n$ the sequence length and $d$ the hidden dimension, and performs attention operation as:
\begin{equation}
\mathrm{Attention}(\queries, \keys, \values) = \mathrm{softmax}\left(\frac{\queries \keys^{\top}}{\sqrt{d}}\right)\values.
\end{equation}

Layer memory modifies key-value generation while retaining the standard query projection.
For the $l$-th layer, queries $\queries$ are obtained from $\hiddenstates_{l-1}$,
but keys and values are derived from $\hat{\hiddenstates}_{l-1}$, a learned fusion of all previous hidden representations ${\hiddenstates_0, \dots, \hiddenstates_{l-1}}$,
where $\hiddenstates_0$ denotes the input embeddings.
This extension allows attention computation to exploit the deeper contextual structure accumulated through the network hierarchy.

\paragraph{Inter-Layer Dynamic Router.}
Inspired by LIMe~\citep{lime}, we introduce an \emph{inter-layer dynamic router} to adaptively weight prior representations, as shown in Figure~\ref{fig:transformer}.
Each layer $l \geq 2$ maintains a router consisting of a learnable linear layer that outputs a context-dependent weighting matrix $\mathbf{R}_l \in \mathbb{R}^{l}$:
\begin{equation}
\mathbf{R}_l(\hiddenstates_{l-1, t}) = \mathrm{Router}_l(\hiddenstates_{l-1, t}), \quad
\mathrm{Router}_l: \mathbb{R}^{n \times d} \rightarrow \mathbb{R}^{n \times l}.
\end{equation}
Here $\hiddenstates_{l-1, t}$ is $\hiddenstates_{l-1}$ modulated by the time embedding of DIT. By doing so, we get a time-aware dynamic router, suitable for diffusion learning.
The aggregated feature map is then computed as:
\begin{equation}
\hat{\hiddenstates}_{l-1} = \mathrm{softmax}(\mathbf{R}_l(\hiddenstates_{l-1, t})) \cdot \hiddenstates_{0:l-1}.
\end{equation}
This adaptive fusion encourages the model to selectively emphasize more informative layers, improving depth-wise consistency and feature reuse.
\begin{wrapfigure}{R}{0.4\textwidth}
\centering\vspace{-1.0em}
\includegraphics[width=0.99\linewidth]{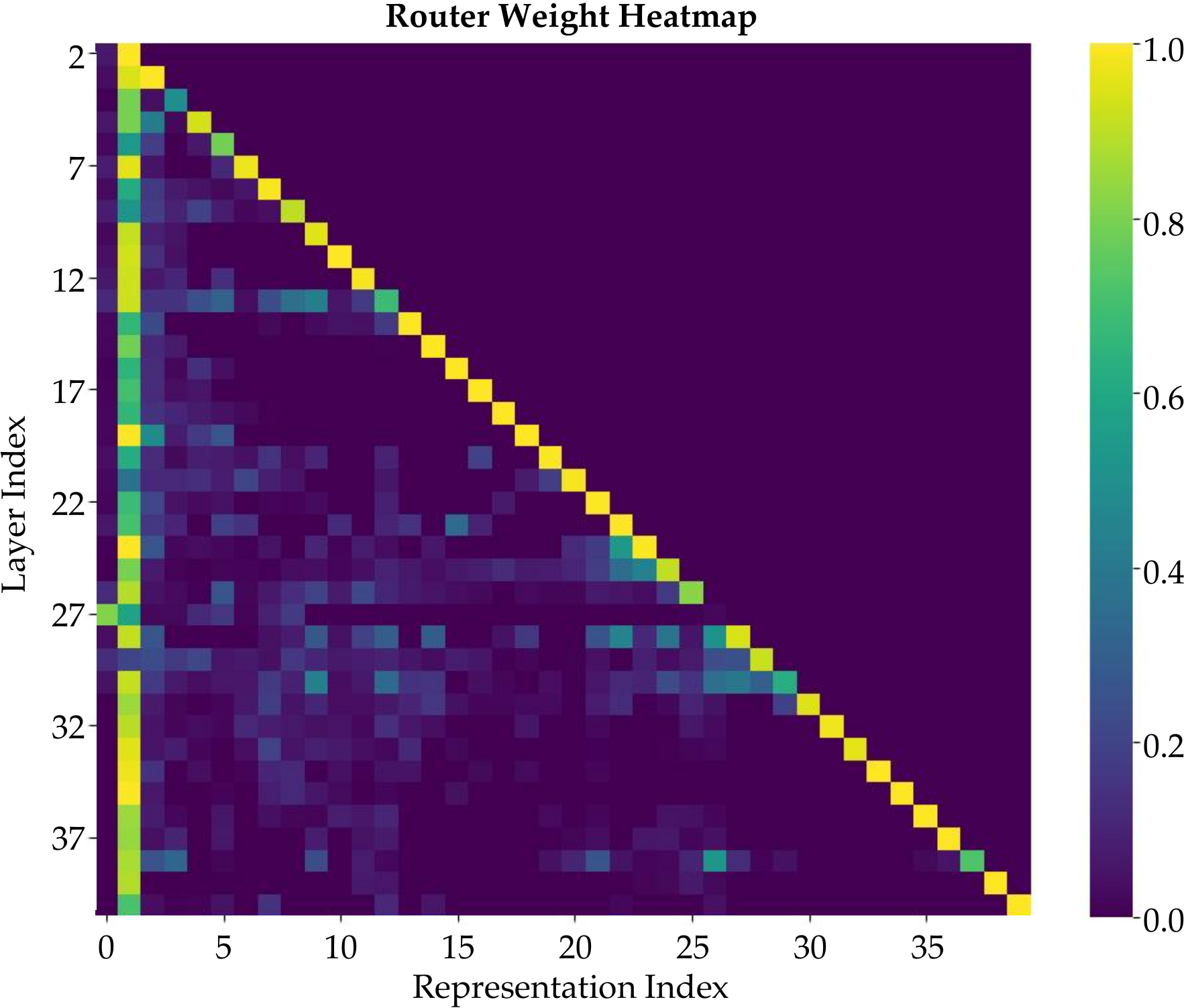} 
\caption{Weight heatmap of the dynamic router. For each representation index in each layer, it shows the maximum value of dynamic router weights across all tokens in a diffusion latent sequence. Layer index starts at 2 because there is no dynamic router for index 0 (input embeddings) and index 1 (only 1 previous layer).}
\vspace{-1.2em}
\label{fig:heatmap}
\end{wrapfigure}

\paragraph{Memory-Aware Self-Attention.}
The self-attention computation at layer $l$ becomes:
\begin{equation}
\begin{aligned}
\queries_{l} &= \hiddenstates_{l-1}\mathbf{W}_{\queries}, \
\keys_{l} &= \hat{\hiddenstates}_{l-1}\mathbf{W}_{\keys}, \
\values_{l} &= \hat{\hiddenstates}_{l-1}\mathbf{W}_{\values},
\end{aligned}
\end{equation}
where $\mathbf{W}_{\queries}, \mathbf{W}_{\keys}, \mathbf{W}_{\values} \in \mathbb{R}^{d \times d}$ are learnable projections.
The query still originates from the previous layer's output, while the keys and values are synthesized through layer-adaptive aggregation.
Notably, our Layer Memory design preserves the original DIT architecture and introduces only minimal additional trainable parameters, facilitating straightforward model adaptation.
The modification remains lightweight and fully compatible with efficient implementations such as FlashAttention~\citep{dao2022flashattention}.
\paragraph{Router Visualization and Analysis.}
The router heatmap in Figure~\ref{fig:heatmap} illustrates how each layer aggregates prior representations for key–value construction. 
A clear diagonal of high weights shows that each layer mainly attends to its immediate predecessor, reflecting the strong sequential dependency typical of standard DITs. In addition, we see a common high weight value for the representation index 1 across all layers. This comes from the router weight for the first latent token, which is mapped to the first frame of the video. This is reasonable, as in image-to-video, the first frame is given as DIT input via channel concatenation~\citep{wan2025wanopenadvancedlargescale}, thus DIT does not need heavy processing of this token in deeper layers, and can borrow information directly from very early layers.
Finally, scattered non-zero weights in the lower-left region indicate that deeper layers selectively reference much earlier representations. 
There is a noticeable cluster of relatively high weights connecting later layers (e.g., layer 13-30) with much earlier representations, indicating that deeper layers are learning to selectively draw low-level, high-frequency features.
The presence of these scattered nonzero weights confirms that the Inter-Layer Dynamic Router is successfully enabling a differentiable memory across depth, allowing the model to selectively bypass intermediate layers to retrieve relevant context directly from any preceding layer when needed.
\paragraph{Training Convergence Analysis.}

\begin{figure}[h]
    \centering\vspace{-1.0em}
    \includegraphics[width=\linewidth]{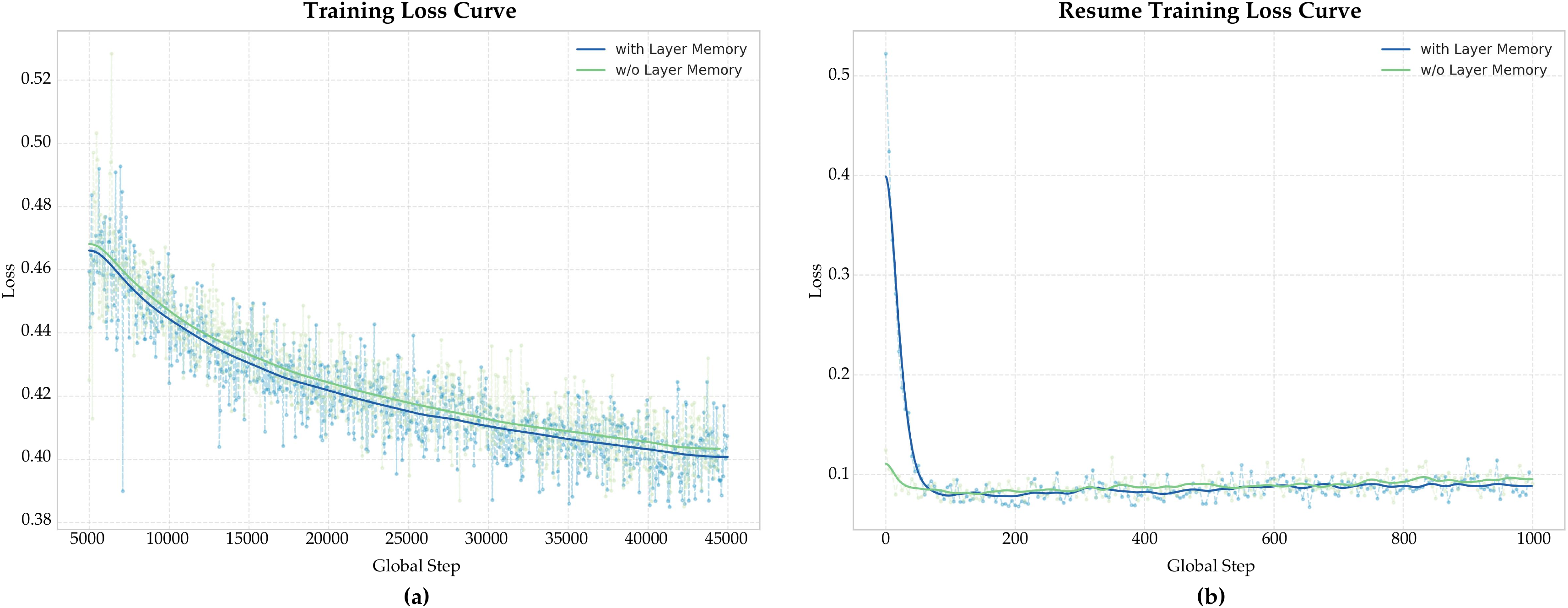}\vspace{-0.5em}
    \caption{Training analysis of the layer memory mechanism.
    (a) Training Loss Curve (From Scratch): Comparison of the training loss between the baseline model (w/o Layer Memory) and the proposed model (with Layer Memory) on Wan2.1-14B-I2V-720P, demonstrating Layer Memory's ability to achieve a consistently lower loss and faster initial convergence. 
    (b) Resume Training Loss Curve (Fine-tuning): Loss curve showing the fine-tuning the original Wan2.1-14B-I2V-720P. The integration of Layer Memory facilitates rapid convergence and sustains a stable performance advantage compared to the baseline. Zoom in to see details.}
    \label{fig:loss_curve}
\end{figure}
The integration of Layer Memory significantly enhances the optimization and convergence of the DIT architecture. As shown in Figure \ref{fig:loss_curve}(a), when training models from scratch, the variant equipped with Layer Memory consistently achieves a lower loss value throughout the entire training process compared to the baseline. This result indicates that the mechanism, by establishing a differentiable memory across depth via the inter-layer router, not only yields a slightly accelerated initial convergence rate but also effectively aids the model in finding a superior minimum on the complex loss surface. 
Furthermore, the Layer Memory mechanism demonstrates its ability to \emph{seamlessly integrate into and improve} existing DiT-structured models. Figure \ref{fig:loss_curve}(b) illustrates the fine-tuning results when Layer Memory is added to a pretrained WAN2.1\citep{wan2025wanopenadvancedlargescale} model. 
The mechanism facilitates rapid convergence within 100 steps and achieves a stable performance gain of up to 4.7\% compared to the baseline after just 1,000 fine-tuning steps. 
This empirical evidence strongly supports the hypothesis that the enhanced inter-layer information flow and adaptive feature reuse improve the model's capacity to learn a more efficient and robust set of representations. We can also integrate the Layer Memory mechanism into MMDIT~\citep{esser2024scaling} architecture for a potentially larger performance gain, since MMDIT has one joint attention across different modalities, hinting at a larger impact of layer memory to each transformer layer. This is left for future work.

\subsubsection{Training Strategy}
\label{subsubsec:train_strat}
\paragraph{Pre-training Stage.}
\input{Sections/2_5_1_pretrain}
\paragraph{Post-training Stage.}
\input{Sections/2_5_2_posttrain}

%% file: Sections/2_5_1_pretrain.tex
We train the DIT using the flow matching framework~\citep{lipman2023flow,esser2024scaling} with logit-normal time sampling schedule. We use Pseudo-Huber loss as the training loss due to its robustness to outliers and lower gradient variance~\citep{song2024improved, Lee2024ImprovedFlow}, making the training more stable. The pretraining data are captioned by Qwen2.5-VL-7B~\citep{bai2025qwen2}, and using a fixed video frame length of 121 frames with 24 FPS. The full pre-training stage is separated into 3 stages. In stage 1, we only use $256\times256$ images to learn text-visual alignment. In stage 2, we switch to $256\times256$ video data to learn video movement and low resolution visual appearance, and finally in stage 3 we use $512\times512$ video data to reach our target resolution. The training step ratio of these 3 stages is approximately $1:2:2$.

%% file: Sections/2_5_2_posttrain.tex
The DIT post-training was separated into the supervised fine-tuning (SFT) stage and the reinforcement learning (RL) stage. For the SFT stage, we finetune our DIT with 300k high-quality samples filtered on aesthetic scores and motion amplitude. This stage increases DIT's generated video quality, but the DIT still suffers from motion integrity problems. Therefore, we apply RL to further improve the DIT, discussed below.

For the RL stage, we use the Reward Feedback Learning (ReFL) framework~\citep{xu2023imagereward} for its high effectiveness to improve video generation quality~\citep{gao2025seedance10exploringboundaries}. Due to a lack of human labeling resources, we opt to use open-source reward models (RM) for our RL training, instead of training RMs from scratch. Specifically, we use VideoAlign~\citep{videoalign} as the video reward model and MPS~\citep{zhang2024learning}, which scores individual video frames, as the frame reward model, and combine the loss of these two models as the reward loss. To reduce the high memory pressure from ReFL training, for each generated video, we only send first 61 frames worth of latent to FSAE's decoder and VideoAlign model, and only 10 frames with the lowest MPS scoring to MPS. This significantly reduces the training memory consumption. We perform multi-round finetuning between the DIT and VideoAlign RM with a small amount of new labeling data generated from DIT each round, while keeping MPS fixed. This approach ensures a stable DIT performance improvement without excessive reward hacking.

During the RL process, we find two tricks that are essential for our training. The first trick is the domain adaptation of the video reward model in the first round of RM training. To do this, we first generate videos using our SFT model, as well as various public models(e.g. ~\citep{wan2025wanopenadvancedlargescale,kong2024hunyuanvideo,yang2025cogvideox}) using the same input prompt and first video frame. Then we combine these generations with ground-truth video to form a candidate pool for each input image plus prompt, and randomly select two candidates as labeling pairs, and finetune VideoAlign on the labeled results. This ensures a smooth domain adaptation from the pre-trained VideoAlign model to our model's data manifold. Experiments show that doing so ensures smooth reward loss increase and model improvement, while not doing this adaptation leads to training divergence. The second trick is the integration of the input image in the reward model input. Here, we finetune the VideoAlign model so that it takes the first frame of the video as an additional model condition when evaluating video quality. Furthermore, in the VLM prompts, we placed emphasis on maintaining consistency between the first frame and the subsequent frames in aspects of color, ID, and details. This aligns with our image-to-video training and significantly improves video consistency.

%% file: Sections/2_4_video_upsampler.tex
\subsubsection{Latent Upsampler}
\label{subsec:latent_upsample}
\begin{wrapfigure}{R}{0.5\textwidth}
\centering\vspace{-1.0em}
\includegraphics[width=0.99\linewidth]{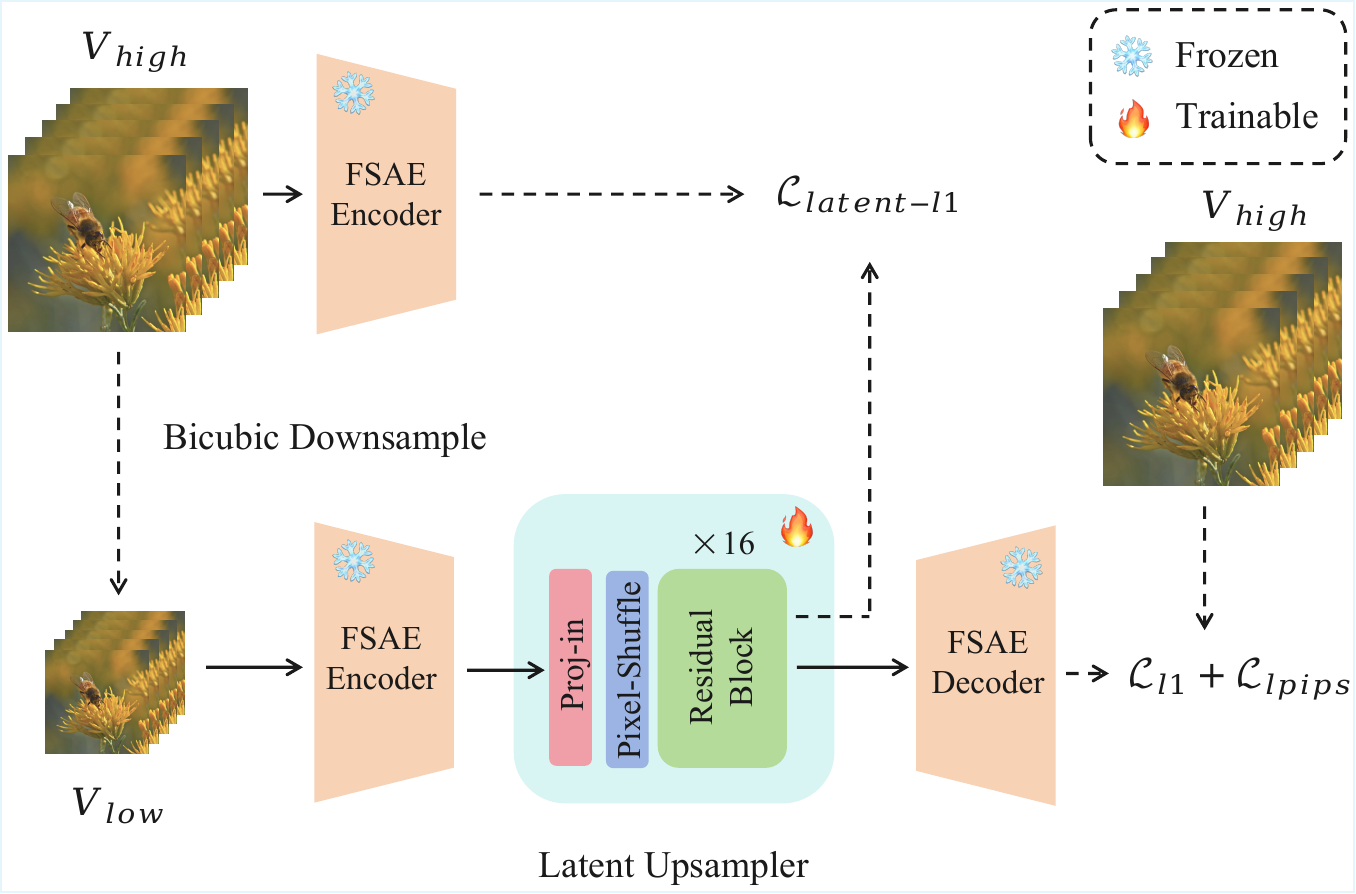} 
\caption{Overall framework of latent upsampler.}
\vspace{-1.2em}
\label{fig:latent_upsampler}
\end{wrapfigure}
After base DIT training, the generated video still suffers from low video detail problems due to the high spatial compression of FSAE. Thus, we propose an additional upsample step to increase video fidelity. The upsample step consists of two modules: (i) a convolutional \textit{latent upsampler}, and (ii) a high resolution DIT \textit{refiner}.  Specifically, the \textit{latent upsampler} module is responsible for upsampling low-resolution latents to the high-resolution feature space, producing a preliminary draft latent. 
The high-resolution \textit{refiner} then conditions on this draft latent to generate the full video sequence. This process preserves the overall structure of the low-resolution input while refining visual details and restoring high-frequency features, resulting in high-quality video output.

For the upsampler module, some work~\citep{polyak2024movie,zhang2025Waver} performs spatial interpolation in RGB space, resulting in extra VAE encoding-decoding operations. Instead, we choose to train a video latent upsampler to upsample low-resolution latents by a factor of 2. Inspired by LTX-Video~\citep{hacohen2024ltx}, we design a convolution-based upsampler starting with a projection layer, followed by pixel-shuffle for upsampling, and then 16 residual blocks. Setting pixel-shuffle before residual blocks results in better upsample quality at the cost of slightly slower speed, which is negligible in the whole FSVideo framework.

We show the latent upsampler training framework in Figure~\ref{fig:latent_upsampler}. The original video $V_{\text{high}}$ is treated as the high-resolution ground truth. We perform bicubic downsample on $V_{\text{high}}$ to get $V_{\text{low}}$. These two videos inputs are encoded separately by the encoder to obtain $Z_{\text{high}}$ and $Z_{\text{low}}$. The low-resolution latent $Z_{\text{low}}$ is fed into our video latent upsampler to generate the upsampled result $\hat{Z}_{\text{high}}$. We then compute the latent L1 loss $\mathcal{L}_{latent-l1}$ between $\hat{Z}_{\text{high}}$ and $Z_{\text{high}}$, and $\mathcal{L}_{l1}$ and $\mathcal{L}_{lpips}$ between $\hat{Z}_{\text{high}}$'s decoder output and $V_{\text{high}}$. The final loss is 
\begin{equation}
    \mathcal{L}_{upsampler} = \alpha_{1}\mathcal{L}_{latent-l1} + \alpha_{2}\mathcal{L}_{l1} + \alpha_{3}\mathcal{L}_{lpips}.
\end{equation}
During training, we set $\alpha_{1}=0.1,\alpha_{2}=0.1$, and gradually increase $\alpha_{3}$ from 0.1 to 1. We also gradually increase the size of the training video from 256p to 1024p and apply the mixed resolution strategy of the VAE training to reduce the memory burden, enabling the latent upsampler to generate latents for 121 frame $1024\times1024$ videos.

\subsubsection{High-resolution Refiner}
\label{subsec:high_res_refiner}
\begin{figure}[t!]
    \centering\vspace{-1.0em}
    \includegraphics[width=\linewidth]{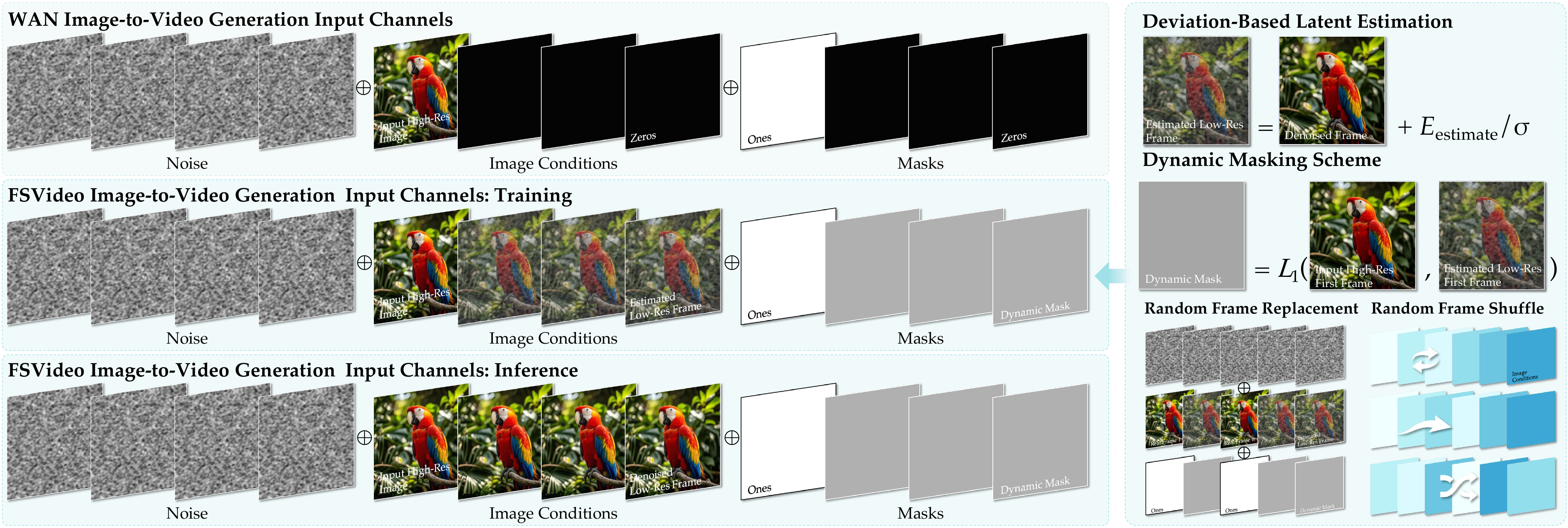}\vspace{-0.5em}
    \caption{
    Enhanced training strategies for the high-resolution video refiner. 
    The figure illustrates three input conditioning channels (\textit{noise}, \textit{image conditions}, \textit{masks}) and our proposed training design. 
    Top: WAN Baseline. Only the first frame is conditioned. 
    Middle: FSVideo refiner training. Incorporates low-resolution video latents and three alignment mechanisms: Deviation-Based Latent Estimation, Dynamic Masking (right), and Random Frame Replacement/Shuffle (bottom right) to improve restoration and temporal robustness. 
    Bottom: FSVideo refiner inference. The refiner performs Video-to-Video generation using the latent upsampler output as its primary condition.} 
    \label{fig:condition}
\end{figure}
We design our video \emph{refiner} model based on FSVideo's base DIT, and one question is how to integrate the latent output from the previous latent upsampler to the refiner. As shown in Figure~\ref{fig:condition}, the common image-to-video DITs, e.g., Wan, extend the input channels to incorporate three conditioning components: \textit{noise}, \textit{mask}, and \textit{condition}.
The \textit{mask} of the first frame is set to $1$ and all subsequent frames to $0$. Similarly, the \textit{condition latent} of the first frame corresponds to the real image, while the remaining frames are set to zero. 
The model performs denoising generation based on the noise component, while the combination of mask and condition provides visual guidance. 
Our refiner follows this general architecture, but is different in the condition setting.

Unlike typical video super-resolution or image-to-video methods, our approach operates in a video-to-video fashion, where the high-resolution first-frame image and subsequent low-resolution frames jointly serve as conditional inputs. This design introduces additional challenges in conditioning alignment and model adaptation:
(1) The refiner is required to ensure faithful adherence to the high-resolution latent while enhancing the low-resolution latent. 
(2) Since the low-resolution inputs are generated by the base DIT and latent upsampler, they may contain artifacts. The refiner is required to perform strong restoration, rather than simply upscaling the inputs. 
(3) As the refiner simultaneously performs detail restoration, it also needs to preserve its text-to-video generation capability. 
To address these challenges, we introduce an series of enhanced training methods, including \emph{dynamic masking, deviation estimation, and conditional dropout}.
\paragraph{Dynamic Masking Scheme.}
\label{subsec:dynamic_mask}
When the task shifts from image-to-video to video-to-video, a straightforward solution is to inject low-resolution latent (upscaled by the latent upsampler, but still of low quality) as condition latents and set all masks to $1$.
However, when both first-frame preservation and subsequent-frame refinement are required, setting all mask values to $1$ and composing the condition latents from both the first-frame and low-resolution latents leads to noticeable visual quality inconsistency across frames. This is because the model struggles to distinguish between genuine latents and low-resolution latents under uniform masking.

To mitigate this confusion, we introduce an \emph{dynamic masking} strategy that adaptively differentiates between real and generated latents. Specifically, given a low-resolution \textsl{first frame} latent $\latent$ and its upsampled output $\hat{\latent}$ from the latent upsampler, we compute their difference $| \latent - \hat{\latent} |$ to estimate the upsampling error for the current sample. This error is then normalized to a predefined range narrower than $[0, 1]$, and assigned as the mask value for the low-resolution latent regions, while the first frame's mask value remains $1$. In this way, we change the mask value from a 0/1 binary to a confidence score between the ground truth and generated latents, guiding the model to refine visual fidelity where necessary.
We also introduce a regularization strategy by randomly replacing a subset of video frames with high-resolution real frames and setting the corresponding masks to $1$, therefore reinforcing the mask cue in all frames.

\paragraph{Deviation-Based Latent Estimation.}
\label{subsec:deviation_estimation}
To further encourage the refiner model to improve on low-resolution latents, we introduce a \emph{deviation-based estimation mechanism} consistent with the flow-matching formulation of diffusion models. 
Recall that in flow matching, the latent state at noise level $\sigma$ is defined as a deterministic interpolation between the clean latent $\latent_0$ and Gaussian noise $\epsilon$:
\begin{equation}
    \latent_\sigma = (1 - \sigma)\latent_0 + \sigma \epsilon,
\end{equation}
where $\sigma \in [0, 1]$ controls the noise intensity. The model is trained to predict the velocity field
\begin{equation}
    \velocity_\sigma = \frac{d\latent_\sigma}{d\sigma} = \epsilon - \latent_0,
\end{equation}
which describes the instantaneous flow direction from the data distribution toward the noise distribution. 
During refiner training, we generated the predicted low-resolution velocity $\hat{\velocity}_\sigma$ from the base DIT for a random $\sigma$. The predicted low-resolution clean latent can then be written as
\begin{equation}
    \hat{\latent_0} = \latent_\sigma - \sigma \hat{\velocity}_\sigma.
\end{equation}

However, using $\hat{\latent_0}$ directly as a low-resolution condition often leads the refiner to overfit the input, neglecting fine-grained restoration. 
To avoid this, we define a deliberately perturbed latent $\tilde{\latent_0}$ as 
\begin{equation}
\begin{aligned}
    \tilde{\latent_0} &= \epsilon - \hat{\velocity}_\sigma = \hat{\latent_0} + ( \epsilon - \hat{\epsilon}) = \frac{\hat{\latent_{0}} - \latent_0}{\sigma} + \latent_0,
\end{aligned}
\end{equation}
where $\hat{\epsilon} = \hat{\velocity}_\sigma + \hat{\latent_{0}}$ is the predicted noise.
When $\sigma$ is close to 1, $\tilde{\latent_0}$ is closer to $\hat{\latent_{0}}$, which is predicted by the base DIT with a latent input of high noise level, meaning $\hat{\latent_{0}}$ is now naturally inaccurate. When $\sigma$ is close to 0, $\tilde{\latent_0}$ receives an error offset $\hat{\latent_{0}} - \latent_0$ amplified by $1/\sigma$. Thus, $\tilde{\latent_0}$ maintains a controlled offset from the true latent across all timesteps, ensuring that the conditional input remains imperfect. 
Consequently, during training, the refiner learns to correct the artifacts rather than simply replicating the low-resolution input, improving its capacity for structural and textural refinement.
During inference, we do not apply this latent deviation technique, and feed the low-resolution latent produced by the first-stage generative model directly into the denoising process as a latent condition, together with the dynamic mask.

\paragraph{Condition Dropout and Frame-Shuffle Strategies.}
\label{subsec:condition_dropout}

To preserve the text-to-video generation capability of the model, we introduce a \emph{condition dropout} strategy, where during training the model randomly omits the low-resolution latents.
We also incorporate a \emph{frame-shuffle} strategy that, for the low-resolution latent condition, with a 50\% probability, we randomly permute local adjacent frames, two nonadjacent frames, or the entire clip in a 6:3:1 ratio, thus synthetically emulating various temporal degradations.

\subsubsection{Refiner Training Strategy}

For high-resolution training, we used the same DiT pre-training strategy (Section~\ref{subsubsec:train_strat}) and initialized the model by inheriting parameters from the base FSVideo DIT, but this time we train on high-resolution $1024\times1024$ video data. During training, we use base FSVideo DIT to generate low-resolution latent, passing it to the latent upsampler, and combine the upsampler output with ground-truth latent of the high-resolution first video frame as the input conditions to the refiner.

To achieve better efficiency for refiner RL training, we perform a light-weight distillation to reduce the refiner's inference cost to 8 network forward evaluation (NFE). We first do CFG distillation, then perform step distillation through progressive distillation~\citep{salimans2022progressive} to 32 steps, followed by SiDA~\citep{zhou2025adversarial} to 8 steps.
This approach reduces inference time by 87\% while maintaining high visual quality, and reducing the NFE to 8 instead of an even lower value gives enough flexibility of the model for RL training.

In RL training, we utilize GRPO~\cite{liu2025flowgrpotrainingflowmatching} since it does not need gradient back-propagation through the FSAE decoder, requiring much less GPU memory compared with ReFL.
Specifically, we adopt the MixGRPO sliding window strategy~\citep{mixgrpo} to reduce search space, which allows for faster convergence, and we use the fine-tuned VideoAlign~\citep{videoalign} during the base DIT RL training stage as the reward model.

%% file: Sections/3_1_implementation_details.tex
All model trainings are implemented using PyTorch 2~\citep{ansel2024pytorch} using Fully Sharded Data Parallel~\citep{zhao2023pytorch} and gradient checkpointing~\cite{chen2016training}. For DIT training, we also apply context parallelization to further reduce GPU memory consumption. We use AdamW~\citep{loshchilov2018decoupled, kingma2014adam} with $\text{betas}=[0.9, 0.95]$ as the optimizer, while changing the learning rate depending on the task.

%% file: Sections/3_2_evaluation.tex
\begin{figure}[t!]
    \centering\vspace{-1.0em}
    \includegraphics[width=\linewidth]{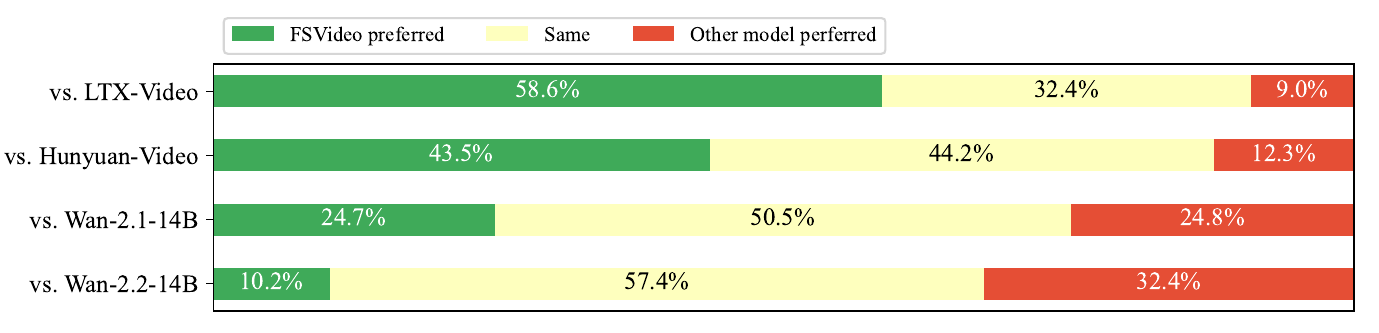}\vspace{-0.5em}
    \caption{GSB evaluation against other models.} 
    \label{fig:gsb}
\end{figure}

\input{tables/vbench_score}
\paragraph{Generation Quality.} To evaluate FSVideo against other leading I2V models, we perform image-to-video evaluation on the VBench 2.0~\citep{zheng2025vbench} I2V benchmark. All experiments adhere strictly to the official pipeline provided by the VBench team, which evaluates models across two core dimensions: I2V Score and Quality Score. Each dimension is further composed of specific sub-evaluation criteria, such as subject consistency, aesthetic quality, motion smoothness, etc. Table~\ref{tab:vbench_score} presents the Vbench results at 720×1280 resolution, with scores from other open source models reported on the official Vbench website. It is clear that FSVideo achieves a competitive score compared to other open-source competitors, only slightly below Step-Video-TI2V, while having the highest video compression rate. Specifically, FSVideo has the best total score compared to other models based on Wan 2.1 DIT~\citep{wan2025wanopenadvancedlargescale, liu2025pusa, chen2025dc}, while having a higher compression rate than other deep compression methods, such as DC-VideoGen.

We also performed a series of human evaluations of FSVideo against other popular open-source video generation models, and show the result in Figure~\ref{fig:gsb}. FSVideo drastically outperforms HunyuanVideo and LTX-Video, and is on par with Wan 2.1 14B, while having a much higher compression rate, resulting in a much faster inference speed (see the \textsl{Inference Speed} part below). Compared to Wan 2.2 14B\footnote{Wan 2.2 is actually a 28B model, considering its MOE design where the diffusion process is separated into two 14B models by different noise level.}, which is the highest-ranked open source image-to-video model in LMArena~\citep{lmarena} and Artificial Analysis Arena~\citep{aa_arena} as of Oct. 2025, FSVideo is less preferred. But given that FSVideo is undertrained given the limited training data and compute. We believe FSVideo's performance can improve even more given more, higher-quality training data, and longer training time.
\input{tables/speed}
\paragraph{Inference Speed.} We evaluate inference speed on H100 GPUs with FlashAttention 3~\citep{shah2024flashattention} installed. For speed evaluation, we mainly look at the DIT inference speed, which is the majority part of the inference computation. Since FSVideo contains two 14B DITs, making it hard to fit both DITs into a single 80G H100, we calculate the inference speed for two cases: 1.) single GPU case where we perform parameter offloading for FSVideo's two DITs, and 2.) duo-GPU case where we can utilize FSDP and context parallelization to reduce memory usage, avoiding parameter offloading. For a comparison, we calculate the speed of Wan2.1-I2V-14B-720P model, and also utilize FSDP and context parallelization for the duo-GPU case for a fair comparison. Evaluations are done for the generation of 5s, $720\times1280$, 24 fps video using BFloat16 precision. 

Table~\ref{tab:speed} shows the evaluation result. We can see that for single-GPU scenario, Wan cannot generate 5 second 24fps even with parameter offloading, while FSVideo succeed with 76.6s. For duo-GPU case, FSVideo achieves a speedup of $\mathbf{42.3\times}$ ,which is much faster than Wan2.1 14B given a similar amount of NFE. If we manage to avoid the GPU memory constraint, such as using FP8 quantization, we will achieve an estimated speed up of at least $\mathbf{58.7\times}$ over Wan2.1 14B. Moreover, we can introduce other speed-up methods such as caching or aggressive step distillation. Since these methods reduce the number of NFEs while our model reduces the computation per NFE, we can achieve a multiplicative speed improvement, further strengthening FSVideo's speed advantage.

%% file: tables/vbench_score.tex
\begin{table}[t]

\centering
\caption{Image-to-Video Generation Results on VBench 720×1280}
\resizebox{\textwidth}{!}{
    \begin{tabular}{l|c|c|c|c|c}
    \toprule
    \multirow{2}{*}{\textbf{Method}} & \multirow{2}{*}{\textbf{Video Autoencoder}} & \multirow{2}{*}{\textbf{\#Params (B)}} & \multicolumn{3}{c}{\textbf{Score}} \\ & & & 
    Total Score$\uparrow$ & I2V Score$\uparrow$ & Quality Score$\uparrow$ \\
    \midrule
    HunyuanVideo-I2V~\citep{kong2024hunyuanvideo} & - & 13 & 86.82\% & 95.10\% & 78.54\% \\
    Step-Video-TI2V~\citep{ma2025stepvideot2vtechnicalreportpractice} & - & 30 & 88.36\% & 95.50\% & 81.22\% \\
    Wan2.1-I2V-14B-720P~\citep{wan2025wanopenadvancedlargescale} & Wan-2.1-VAE-f8t4c16 & 14 & 86.86\% & 92.90\% & 80.82\% \\
    Pusa-V1.0~\citep{liu2025pusa} & Wan-2.1-VAE-f8t4c16 & 14 & 87.32\% & 94.84\% & 79.80\% \\
    DC-VideoGen-Wan-2.1-14B~\citep{chen2025dc} & DC-AE-V-f32t4c32 & 14 & 87.73\% & 94.08\% & 81.39\% \\
    \rowcolor{blue!10} FSVideo & FSAE-f64t4c128 & 14+14 & 88.12\% & 95.39\% & 80.85\% \\
    \bottomrule
    \end{tabular}
}
\label{tab:vbench_score}
\end{table}

%% file: tables/speed.tex
\begin{table}[t]
\centering
\caption{DIT inference speed comparison of 5 seconds $720\times1280$ 24 fps video generation.}
\resizebox{\textwidth}{!}{
    \begin{threeparttable}
        \begin{tabular}{c|c|c|c}
        \toprule
        \multirow{3}{*}{\textbf{Number of H100 GPUs}} & \multicolumn{2}{c}{\textbf{Latency (in seconds, BFloat16 precision)}} & \multirow{3}{*}{\textbf{Speedup}} \\ & Wan2.1-I2V-14B-720P & FSVideo & \\ & 60 NFE\tnote{*} & 68 NFE (60 base DIT + 8 refiner) & \\
        \midrule
        1 & Out of memory & 76.6s (with parameter offloading) & \cellcolor{blue!10} \\
        2 & 822.1s & 19.4s & \cellcolor{blue!10}$42.3\times$ \\
        \midrule
        1 (if no GPU memory constraint) & 1607.5s\tnote{**} & 27.4s & \cellcolor{blue!10}$58.7\times$\\
        \bottomrule
        \end{tabular}
        \begin{tablenotes}
            \item[*] For Wan and FSVideo base DIT, 2 NFE = 1 diffusion step with classifer-free guidance; for FSVideo refiner, 1 NFE = 1 diffusion step due to CFG distillation.
            \item[**] Estimated using 5 seconds 16 fps video generation speed, which takes 1076.1 seconds.
        \end{tablenotes}
    \end{threeparttable}
}
\label{tab:speed}
\end{table}

%% file: Sections/4_conclusion.tex
We introduce FSVideo, an image-to-video framework for fast video generation. At its core is a new video autoencoder with a high compression ratio for token reduction, a new DIT architecture with layer memory self-attention to improve dit feature information reuse, and a multiscale generation strategy with a carefully designed latent refiner to increase video fidelity. Compared with other video generation models of similar parameter size, FSVideo is able to generate competitive videos while being \emph{a order of magnitude faster}, and its inference speed can be further enhanced through post-training distillation methods such as step distillation and model distillation.

Future work may explore: 1.) better video encoding strategy to support higher resolution generation; 2.) new DIT design to further reduce generation time and increase generation quality without significantly increasing model size; 3.) improving training methods for better prompt coherence and larger video movement, 4.) extending to multimodal generation such as video+sound generation or video editing, and 5.) extending to longer videos or multiscene videos. 

Finally, we would like to have a quick discussion on the future of efficient models. Past works have shown that model speed-up techniques by reducing model capacity, either by distillation of the model parameters or by designing lightweight modules, usually come with a trade-off between speed and quality~\citep{efficient_survey} that ultimately produces unsatisfactory results. We need the model capacity to be big enough to model complex training-data distributions, especially for large generative models. We believe that a promising direction is to \emph{reduce the token amount per model inference and increase token efficiency}, which applies to both training and inference. This can be done by looking in the following directions: 1.) A more compact and meaningful token representation space, such as \citep{chen2025deep, rae} for DIT and \citep{wei2025deepseekocr, cheng2025glyph} for language models; 2.) a better model and training design to increase token usage efficiency, such as \citep{fang2023cross,zhu2024hyper,lime} by adding more token passages in transformers, \citep{shazeer2017outrageously, krause2025tread} for adaptive token routing, \citep{roy2025fast} for \textsl{higher} attention computation to improve token efficiency, and to some extent, few-step distillation and training methods~\citep{song2023consistency, geng2025mean}, since they enforce each token to have a high representation capacity for few-step inference; and 3.) a better training optimizer, for example Muon~\citep{jordan2024muon}. We believe video generation can be both high-quality and cost-effective, and we hope our work will help pave this path in the long run.